**DelayPTC-LLM: Metro Passenger Travel Choice Prediction under Train Delays with Large Language Models**

**Chen Chen**
College of Urban Transportation and Logistics
Shenzhen Technology University, Shenzhen, China, 518118
Email: 2310414021@email.szu.edu.cn

**Yuxin He***
College of Urban Transportation and Logistics
Shenzhen Technology University, Shenzhen, China, 518118
Email: heyuxin@sztu.edu.cn

**Hao Wang**
College of Electronics and Information Engineering,
Shenzhen University, Shenzhen, China, 518061
Email: haowang@szu.edu.cn

**Jingjing Chen**
College of Urban Transportation and Logistics
Shenzhen Technology University, Shenzhen, China, 518118
Email: chenjingjing@sztu.edu.cn

**Qin Luo**
College of Urban Transportation and Logistics
Shenzhen Technology University, Shenzhen, China, 518118
Email: luoqin@sztu.edu.cn

Word Count: 5,431 words + 3 table (250 words per table) = 6,181 words

*Chen Chen, Yuxin He\*, Hao Wang, Jingjing Chen and Qin Luo*

## ABSTRACT

Train delays can propagate rapidly throughout the Urban Rail Transit (URT) network under networked operation conditions, posing significant challenges to operational departments. Accurately predicting passenger travel choices under train delays can provide interpretable insights into the redistribution of passenger flow, offering crucial decision support for emergency response and service recovery. However, the diversity of travel choices due to passenger heterogeneity and the sparsity of delay events leads to issues of data sparsity and sample imbalance in the travel choices dataset under metro delays. It is challenging to model this problem using traditional machine learning approaches, which typically rely on large, balanced datasets. Given the strengths of large language models (LLMs) in text processing, understanding, and their capabilities in small-sample and even zero-shot learning, this paper proposes a novel Passenger Travel Choice prediction framework under metro delays with the Large Language Model (DelayPTC-LLM). The well-designed prompting engineering is developed to guide the LLM in making and rationalizing predictions about travel choices, taking into account passenger heterogeneity and features of the delay events. Utilizing real-world data from Shenzhen Metro, including Automated Fare Collection (AFC) data and detailed delay logs, a comparative analysis of DelayPTC-LLM with traditional prediction models demonstrates the superior capability of LLMs in handling complex, sparse datasets commonly encountered under disruption of transportation systems. The results validate the advantages of DelayPTC-LLM in terms of predictive accuracy and its potential to provide actionable insights for big traffic data.

*Chen Chen, Yuxin He\*, Hao Wang, Jingjing Chen and Qin Luo*

## INTRODUCTION

The Urban Rail Transit (URT) system has become an integral component of modern society, significantly enhancing transportation efficiency and alleviating traffic congestion. However, the expansion of network lines and increasing passenger flow have occasionally led to train delays. These delays, caused by factors such as equipment malfunctions and natural disasters, are both occasional and unpredictable, profoundly affecting passengers' daily commuting experiences and diminishing URT service quality. When metro delays inevitably occur, the top priority for operation management is to recover services promptly. To alleviate its negative impacts and improve URT operation management, it is crucial to understand how metro delays affect passengers' travel choice behaviors and predict these choices accurately. This involves leveraging established travel patterns and available delay information to forecast passengers' choices, which in turn facilitates the refinement and enhancement of emergency response strategies. However, predicting passenger behavior under such conditions poses significant challenges. These challenges arise from the dynamic and complex nature of passenger choices, the imbalance and sparsity of data due to the infrequent occurrence of delays, and the diverse factors influencing travel decisions, including delay factors, environmental conditions, and individual preferences.

Previous research predominantly explores macroscopic aspects such as passenger evacuation, traffic control, delay propagation, and adjustments in train operations, often overlooking the microscopic passenger travel choice behavior (*1–6*). Earlier studies on passenger travel choices often rely on random utility theory and regret theory, employing survey data to analyze the decision-making processes (*7–8*). However, such survey data suffers from subjectivity, low accuracy, and respondent biases. Furthermore, theories like random utility theory, while insightful, excessively rely on assumptions of rational behavior, neglecting the heterogeneity among passengers limiting practical application.

With the rapid expansion of transportation big data, data-driven methods have become the predominant research approach due to their objectivity, precise predictability, and adaptability (*9–11*). However, traditional data-driven techniques like statistical analysis, machine learning, and natural language processing (NLP) face limitations in dealing with data quality, model parameters, and selection. These methods often require large, balanced datasets and struggle with sparse and imbalanced data.

Given the advantages of Large Language Models （LLMs）such as GPT-4 (OpenAI, 2023), Claude (Anthropic,2024), and Llama (Touvron,2023), in processing structured data, learning from imbalanced samples, and generating human-readable explanations, we propose a framework for Passenger Travel Choice prediction under metro Delays with LLMs (DelayPTC-LLM). LLMs, built on transformer architectures with extensive parameters, demonstrate exceptional generalization and predictive capabilities. They excel in handling sparse data and providing insights into complex patterns, making them well-suited for predicting passenger travel choices under metro delays (*12–14*).

The overall prediction procedure is divided into the following five steps: data profiling, regular passenger screening, affected regular passenger identification, travel choices dataset construction, and travel choice prediction with DelayPTC-LLM. DelayPTC-LLM comprises three key components:

(1) **Integration of Personal Characteristics and Delay Information**: We utilize AFC data to explore travel patterns and preferences, combining this information with delay event records to predict travel choices under metro delays.

(2) **Utilization of LLMs Instead of Resampling**: Unlike traditional methods that rely on substantial labeled data and oversampling, LLMs handle sparse and imbalanced data, overcoming conventional model limitations.

(3) **Mining Travel Choices and Establishing Prompt Templates**: We process raw AFC data to identify affected passengers and determine their choices by comparing delay day records with normal days. A variety of tailored model commands are established for the dataset, and prompt engineering identifies the model with optimal prediction performance. This paper validates DelayPTC-LLM via real-world datasets of Shenzhen Metro in China.

The principal contributions of this research are as follows:

(1) **Pioneering the Use of LLMs for Delay Condition Passenger Choice Prediction**: This study is the first to employ LLMs to address passenger choice prediction under delay conditions. By considering passenger heterogeneity and the interaction between travel data and delay information, our approach effectively resolves issues related to data imbalance and limited sample size in delay conditions.

(2) **Innovative Chain-of-Thought (CoT) Prompt Engineering**: We develop a series of tailored prompt templates that significantly enhance prediction accuracy and interpretability, enabling the model to understand and predict passenger behavior more effectively.

(3) **Exploration of LLM Potential in Explainable Model Development**: Through comparative experiments, we explore the potential and prospects of LLMs in developing interpretable models for similar tasks, highlighting their applicability and future research directions.

## METHODOLOGY
### Definition and Problem Formulation
*Research scenario and the definition of regular passenger, passenger heterogeneity*
Metro delay refers to the situation where a train deviates from its planned path due to various factors, such as equipment operation, natural disasters, or organizational management issues when traveling according to the train operation diagram.

This research concentrates on regular passengers of urban rail transportation as the subject for further research. Regular passengers are defined as those who exhibit stable travel patterns. To more accurately discern the impact of metro delays on passengers' travel, we separate passengers' travel demand into fixed regular patterns and irregular deviations. Finding fixed travel patterns is challenging with infrequent travelers due to their limited travel records. Similarly, for passengers with high travel frequency but no consistent spatial distribution, identifying patterns in their travel behavior becomes difficult. This study utilizes two criteria—the number of travel days and spatial consistency—with thresholds to select regular passengers.

When confronted with the same metro delay, passengers primarily make their decisions based on the travel costs of various alternative options. However, passengers' travel choices are not entirely consistent, mainly because passengers consider both objective factors and individual preferences. Passengers' heterogeneity directly shapes diverse passenger behavior, significantly affecting their travel choices. Thus, to enhance the model accuracy and align theoretical research with practical scenarios, it is crucial to investigate passenger travel choice behavior under metro delays based on passenger heterogeneity. We need to delve into each passenger's travel choices from a micro-perspective.

In this research, we integrate regular passengers' travel pattern data with information on metro delays to analyze affected regular passengers. Considering passenger heterogeneity, we compare affected regular passengers' stable travel behaviors with behaviors on delay days to mine their travel choices during metro delays. Affected regular passengers' travel choices dataset under metro delay will be utilized as inputs for LLMs in the subsequent travel choice prediction process, as will be introduced in the following section.

*Problem Formulation*
Factors such as the delay type, the trip purpose, the delay period, and the weather have all been identified as having a potential impact on passenger travel choice in the presence of train delays. Passengers' travel decision-making process is intricately complex, with their travel choice behaviors being influenced by a multitude of factors. The metro delay event, alongside passengers' individual preferences, significantly shapes their travel decisions. Critical factors such as the trip purpose, delay period, and prevailing weather conditions have been identified as key determinants influencing passengers' choices in the context of metro delay.

In this research, we identified five principal features: type of delay, time of delay, trip duration, whether the passenger had already started their journey at the time of the delay, and the urgency level of the trip. The first two features, type of delay and time of delay, are directly associated with the delay incident itself. The remaining three features pertain to the travel habits of affected regular passengers. For a given metro delay event v, the related event features are defined as $E_v = \{v_1, v_2\}$ ,indicating the delay type and delay period. For a given affected regular passenger p, the related features are defined as $E_p = \{p_1, p_2, p_3\}$ ,indicating the last three features of the passengers, respectively.

We analyze AFC data, selecting passengers who travel frequently and identifying their travel patterns. To determine whether these passengers are impacted by metro delays, we employ the principle of spatiotemporal overlap. By comparing travel records from delay days to those from normal days, we discern the travel decisions made by passengers in response to metro delays. Utilizing these travel decisions as the target outcome, and incorporating the previously mentioned five factors as predictive features, we establish a passenger travel choice prediction model under metro delays based on the travel choices dataset.

The dataset contains delay event features and passenger features. The methods for obtaining these features are as follows: delay type and delay period are extracted from the delay event logs, the average trip duration and urgency level are derived from passengers' personal historical travel records, and whether passengers commenced their trip during the delay period can be assessed based on their travel records on delay days. This comprehensive approach allows us to gain insights into both the nature of delays and passengers' behavior in response to them. The last three features cannot be described quantitatively. To improve the predictive performance of the model, we need to perform feature engineering. For a regular passenger affected by the delay, average trip duration refers to the average travel time for all trips that belong to their affected travel patterns,which can be inferred from the historical inbound and outbound time of the affected trips. According to the start time of the delay event and the inbound time of the affected trip, we can determine whether the passenger started the trip before the delay event. We define urgency level as the standard deviation of the inbound times for passenger's affected travel patterns. A smaller standard deviation for a particular travel pattern indicates that the trips are more likely to be urgent. Based on this dataset, we develop a travel choice prediction model during delays, based on both passengers' personal attributes and characteristics of delay events. In essence, this model forecasts passengers' responses to delays, leveraging both individual and event-specific attributes.

Our goal is to predict the passenger travel choice under delay event denoted as $Y$ based on historical AFC data and delay event records. This can be formulated as:

$$Y = F(E_v, E_p) \qquad (1)$$

where $F$ represents the prediction model to be developed. Further details will be elaborated upon subsequently.

**Overall Procedure for Passenger Travel Choice Prediction Under Train Delay**

We propose a passenger travel choice prediction framework under metro delays. We compile a list of regular passengers affected by metro delays, this involves analyzing the travel patterns of regular passengers and pinpointing their locations. Upon identifying passengers affected by metro delays, we analyze whether they alter their planned routes, thus understanding their travel choices. Utilizing the travel choices dataset derived from these observations, we construct travel choice prediction models using DelayPTC-LLM. In essence, our framework integrates five core components: data profiling, regular passenger mining, affected regular passenger identification, travel choices dataset building, and passenger behaviors prediction modeling, as depicted in **Figure 1**.

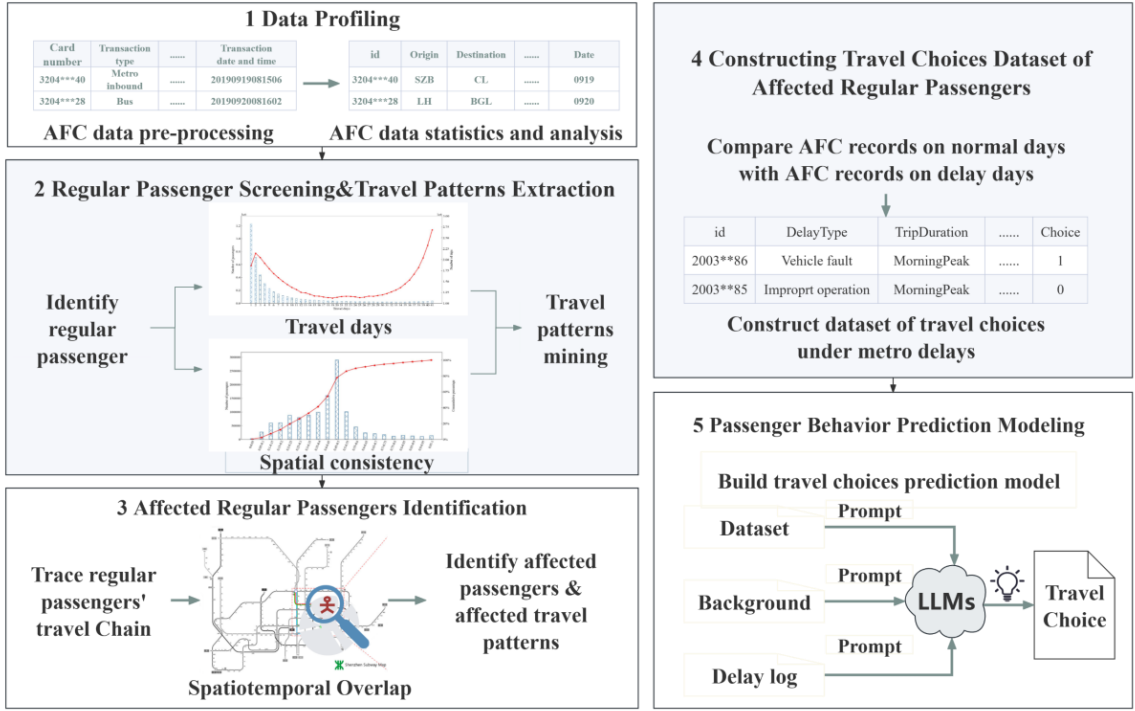

**Figure 1 Overall procedure for passenger travel choice prediction under train delay**

Step 1: Data profiling. We begin by pre-processing the raw AFC data and conducting statistical analysis on the processed AFC data from travel days and spatial consistency to ensure the effectiveness and accuracy of passenger flow patterns.

Step 2: Regular passenger screening and travel patterns extraction. According to processed AFC data, passenger travel patterns are statistically analyzed, and regular passengers are screened out using indicators such as travel days. For regular passengers, according to the spatial and temporal distribution characteristics of their historical trips, we use the clustering method to mine their travel patterns.

Step 3: Affected regular passenger identification. A spatial-temporal overlap approach is proposed to determine affected passengers and their affected travel patterns. Given a specific delay event, we determine whether a passenger is affected by the delay by judging the spatial-temporal overlap degree between the passenger's trip and the delay.

Step 4: Constructing a travel choices dataset of affected regular passengers. Compare the travel records of affected regular passengers under metro delays with their normal travel patterns to obtain the travel choices dataset under metro delays. The travel choice behaviors of affected regular passengers can be divided into the waiting type and abandonment type. Waiting type passengers choose to remain at the station or opt to delay their departure, and then continue their trip by metro as usual. Abandonment type passengers choose to abandon the metro and use an alternative mode of transportation.

Step 5: Establish a passenger travel choice behaviors prediction model under metro delays. Based on the affected regular passenger travel choices dataset, a travel choice prediction model under metro delays is established based on LLMs. The DelayPTC-LLM is proposed to predict whether an affected regular passenger will abandon URT due to a specific metro delay.

## The architecture of DelayPTC-LLM
*Framework Overview*

DelayPTC-LLM consists of two primary stages, as depicted in **Figure 2**: (1) Delay log mining. LLMs employ language processing capabilities to analyze raw delay logs, extracting key details such as causes, impact scope, and time of metro delays. This process transforms complex text logs into clear, structured delay event features. (2) Passenger travel choice prediction. LLMs process the dataset of passenger travel choices under delays by handling missing data, selecting relevant features, and transforming these features. Based on the data features and instructions, it predicts travel choices for the processed dataset. Due to random factors such as differences in data features and prompt words, the processing methods of LLMs are roughly divided into three categories:(i) Statistical methods. After segmenting the dataset, the LLMs train on statistical methods like discriminant analysis and naive Bayes using the training set. It then evaluates and optimizes these models, culminating in the generation of predictions. (ii)Machine learning methods. The dataset is split into training and testing sets. Based on the data's distribution, suitable models like LGBM, DBSCAN, or XGBoost are chosen. The models are trained and evaluated using cross-validation to prevent overfitting. If the results are subpar, hyperparameters are refined via grid or random search to enhance performance. Finally, predictions for each data point's travel choices are produced. (iii)NLP methods.LLM leverages delay logs and travel choice descriptions, converting structured data into narrative forms. Following the encoding of categorical features and the standardization of numerical features, it uses purely NLP methods to predict travel choices under metro delay conditions.

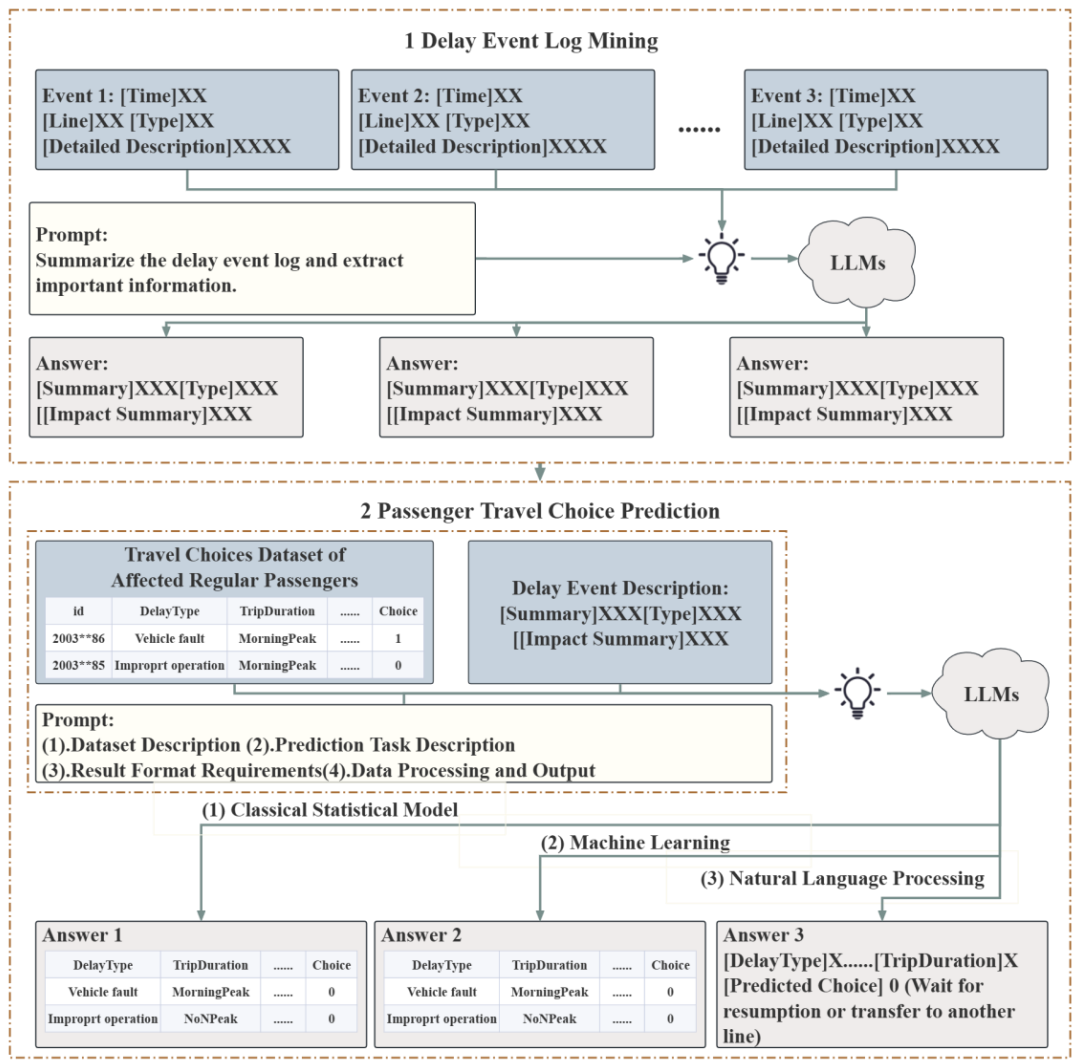

**Figure 2 The framework of DelayPTC-LLM**

*Delay feature formatting*

Delay logs are comprehensive records within URT system that capture details of operational disruptions and fault resolutions. Delay logs include information on the delay type, the time of occurrence, impact, and specific descriptions of each delay event. These logs facilitate analyses of fault frequency, type, and their effects on operations. By feeding these logs into LLMs, the models can uncover underlying information like the extent and frequency of delays, providing summarized insights and characteristic features for each recorded delay. Furthermore, by merging delay logs with passenger behavioral data, LLMs assess factors such as the delay time, the delay type, and duration that may influence passengers' travel choices during metro delay events. This integration enables the development of refined prediction models that anticipate passengers' reactions to specific delay events.

*Passenger choice prediction with prompts*

LLMs can receive inputs in the form of natural language and unprocessed vectorized data. In this research, we utilize LLMs as passenger travel choice predictors, using processed delay event description and passenger travel choices dataset under metro delays as input.

Our model aims to predict passengers' travel choices under delays accurately. To harness the reasoning and few-shot learning capabilities of DelayPTC-LLM, we design the prompt. The prompts begin with detailed instructions detailing task inputs and expected outputs and conclude with guidelines on how to approach the prediction tasks, emphasizing the simultaneous consideration of features and delay event information.

Additionally, we require that DelayPTC-LLM not only generate predicted inflow and outflow values but also clearly describe the reasoning process behind these predictions by directing the model to systematically consider each step before advancing. This approach is referred to as the 'Chain-of - Thought'. CoT prompts have demonstrated significant capabilities in executing complex reasoning tasks. Integrating CoT prompts is deemed essential for guiding the model to deeply consider aspects such as passenger heterogeneity, delay propagation, and the spatial dynamics within the metro network. We design CoT prompts by breaking down the task into three sub-questions that guide the model through the reasoning process, facilitating the solution to the ultimate prediction problem.

**Figure 3** illustrates an example of the prompt, consisting of dataset description, prediction task description, CoT, and so on. It defines the DelayPTC-LLM' role within the operational management department of a URT system, where it is tasked with predicting passenger movements during delays, using historical travel records and delay data. Moreover, if NLP methods are employed, the original dataset must be converted into a textual format. These prompts can then instrucrt DelayPTC-LLM' to automatically produce coherent text.

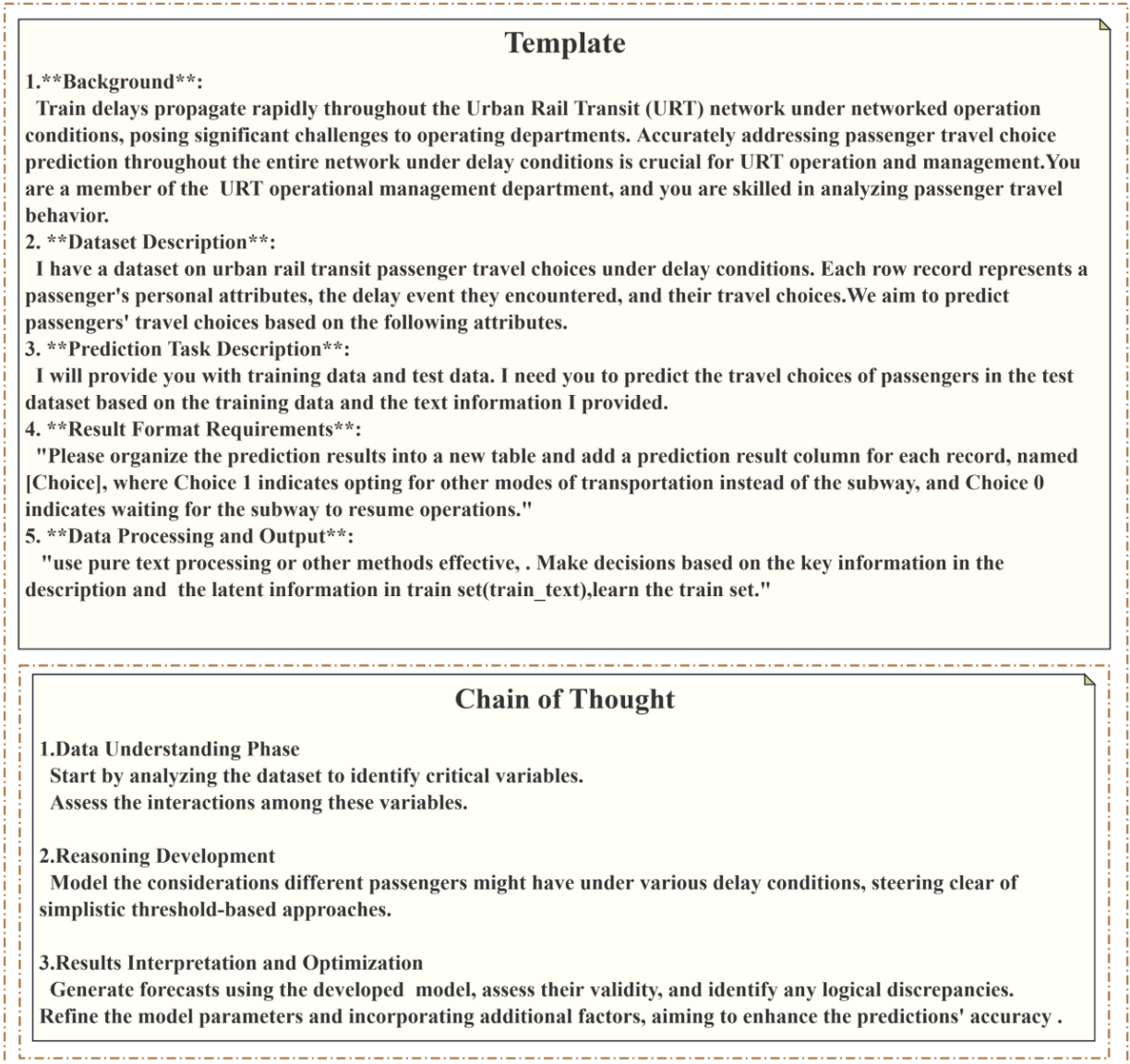

**Figure 3 The prompt template of DelayPTC-LLM**

## CASE STUDY
### Data Description
*AFC Data*

The AFC system is a comprehensive automated framework that manages ticket sales, inspections, fare collection, financial settlements across various entities, and overall operation management. The extensive and reliable data generated by the AFC system are readily accessible and provide significant advantages over traditional manual surveys. Such data markedly improves the capability of operational management to analyze and understand the travel needs and behaviors of passengers, thereby offering crucial support for the intelligent operation management of the URT system.

The Shenzhen Metro AFC data utilized in this study were provided by Shenzhen Metro, China. Notably, comprehensive AFC data typically encompasses Shenzhen Tong cards, single-journey tickets, day passes, and QR code-based ticketing. However, due to limitations in data availability, this research only incorporated data from Shenzhen Tong.

The AFC data encompasses all transactions made with Shenzhen Tong cards from August 1 to September 30, 2019, excluding August 4, August 26, September 15, and September 29. It includes data from 41 weekdays and 18 holidays. Due to significant differences in travel choice behaviors between weekdays and holidays among urban rail passengers, this study primarily analyzes the weekday travel data. The methodologies and framework proposed can be applied to weekends and holidays, enhancing their broad applicability.

AFC data consists of complete information about swipe card records for inbound and outbound at the station. Each data record represents a swipe transaction of a passenger and includes details such as the metro station or bus route, swipe time, transaction type, transaction amount, device code, license plate number, etc. Smart cards are assigned unique card numbers, facilitating the identification of individual passengers.

**Table 1** presents a sample of the original Shenzhen Metro AFC data.

**TABLE 1 Example of Shenzhen Metro AFC Data**

| ID | Transaction Date and Time | Transaction Type | Company | Line/Station |
|---|---|---|---|---|
| 6878***27 | 20190915190008 | Metro (Entry) | Line 5 | Changlong |
| 3204***40 | 20190915185759 | Metro (Exit) | Line 5 | Yangmei |
| 3230***96 | 20190915200725 | Bus | Bus Group | M429 |
| 022*****98 | 20190914164700 | Bus QR Code | Eastern Bus | M429 |
| 022*****09 | 20190914164751 | Bus QR Code | Western Bus | 802 |

*Delay Data*

Shenzhen Metro suffered a total of 14 train delays from August 1, 2019, to September 30, 2019, which were used to train the prediction model of the affected regular passengers' choices. The original delay logs contain specific delay information such as the line, fault type, delay time, specifics of the fault, and the segments affected. An example of the delay event log is illustrated in **Figure 4**. After LLM's analysis, the refined key information is presented in **Table 2**.

Delay Event Log

Event : On August 20, 2019, at 7:50 AM, the driver of train 03503 (551) reported an object jamming the door on the down line at Minzhi, causing the door to neither open nor close. The driver proceeded to the scene for resolution. By 7:55 AM, the door malfunction had been resolved. Due to this incident, train 03703 (241) and train 03903 (513) destined for Chiwan were delayed by 4 minutes 24 seconds and 6 minutes 21 seconds, respectively. Additionally, train 03503 (551) and train 00605 (510) destined for Qianhaiwan experienced delays of 8 minutes 55 seconds and 8 minutes 42 seconds, respectively. On the following day, at 6:25 AM, further inspections were conducted on the train, confirming no recurring issues.

**Figure 4 An example of delay event logs**

**TABLE 2 Delay Event Information**

| Line | Delay type | No. | Date | Time | Delay interval | Direction |
|------|-----------|-----|------|------|----------------|-----------|
| Line 1 | Vehicle Fault | 1 | 2019-08-27 | 08:10-09:09 | Taoyuan-Luohu | Up |
| | | 2 | 2019-09-26 | 08:57-09:49 | Pingzhou-Airport East | Down |
| | Signaling Fault | 3 | 2019-09-19 | 18:04-19:08 | Shenzhen University-Airport East | Down |
| | | 4 | 2019-09-20 | 16:03-17:10 | Xin'an-Qianhaiwan | Up |
| | Power Fault | 5 | 2019-09-26 | 06:31-06:55 | Grand Theater-Luohu | Up |
| | | 6 | 2019-09-26 | 06:31-07:50 | Luohu-Airport East | Down |
| Line 5 | Vehicle Fault | 7 | 2019-08-26 | 07:49-08:52 | University Town-Huangbei Ling | Up |
| | | 8 | 2019-09-30 | 07:53-09:13 | Xiashuijing- Qianwan Park | Down |
| | Improper Operation | 9 | 2019-08-20 | 07:54-09:14 | Bao'an Center-Huangbei Ling | Up |
| | Others | 10 | 2019-08-01 | 09:32-10:41 | Baigelong- Qianwan Park | Down |
| | | 11 | 2019-08-07 | 07:55-08:36 | Tanglang- Qianwan Park | Down |
| | | 12 | 2019-08-20 | 07:50-08:36 | Minzhi- Qianwan Park | Down |
| Line 11 | Power Fault | 13 | 2019-08-12 | 08:08-09:08 | Fuyong-Futian | Up |
| | Others | 14 | 2019-08-28 | 08:22-09:33 | Bihaiwan -Bitou | Up |

**Experiment Settings**

In our experiments, we employed OpenAI's advanced LLM, GPT-4. The model is a deep learning-based natural language generator trained on extensive text corpora, capable of producing text that is both coherent and contextually relevant. Utilizing the Transformer architecture, which is powered by an attention mechanism, GPT-4 efficiently manages long-range dependencies within the text, excelling in complex text comprehension and the generation of smooth, coherent language. The training leverages a broad range of text data, including books, articles, web pages, and more. GPT-4 has shown outstanding performance across various language processing tasks like text summarization, question answering, dialogue generation, and text classification. Additionally, it has the capability to directly receive, analyze, and process user-provided datasets, conducting tasks such as data cleaning and transformation, thus showcasing robust responsiveness.

In this research, the DelayPTC-LLM is configured to utilize GPT-4. We aim to assess how GPT-4 performs in predicting passenger travel choice behavior under metro delay conditions. The passenger travel choices dataset under metro delays includes details like the type of delay and the duration of passenger travel.

To assess the performance of passenger travel choice prediction, we compared our proposed DelayPTC-LLM method with the efficiency of pure machine learning methods, GPT-4, machine learning combined with GPT-4:

i. GPT-4: Provide basic background information and output requirements only, GPT-4 autonomously processes the dataset.

ii. GPT-4o: GPT-4 Optimal, is an optimized version of OpenAI's GPT-4 model, designed for enhanced efficiency and faster response times while maintaining advanced natural language understanding and generation capabilities. Provide basic background information and output requirements only, GPT-4o autonomously processes the dataset.

iii. RF: Random Forest is an ensemble learning method that constructs a multitude of decision trees at training time and outputs the class that is the majority vote of the individual trees, offering robustness against overfitting by averaging multiple deep decision trees trained on different parts of the same training set.

iv. LGBM: LightGBM is a fast, distributed, high-performance gradient boosting framework based on decision tree algorithms, used for ranking, classification, and many other machine learning tasks, which is efficient in handling large-scale data and focuses on the accuracy of results.

v. RF +GPT-4: Instruct GPT-4 to assist in the preparation or analysis of data for training an RF model.

vi. LGBM +GPT-4: Instruct GPT-4 to assist in the preparation or analysis of data for training a LGBM model.

We conducted the analysis using four evaluation metrics:

i. Accuracy: The proportion of true results (both true positives and true negatives) among the total number of cases examined.

ii. Recall: Also known as sensitivity or true positive rate, it measures the proportion of actual positives that are correctly identified by the model. It is calculated as the ratio of true positives to the sum of true positives and false negatives.

iii. Precision: The proportion of positive identifications that were correct. It is calculated as the ratio of true positives to the sum of true positives and false positives.

iv. F1 Score: The F1 Score is the harmonic mean of precision and recall, and it is used to balance the two metrics, especially when the class distribution is uneven.

## RESULT ANALYSIS

### Performance Comparison with Baseline Models

**Table 3** summarizes the passenger travel choice prediction performance of different models on the dataset. From the experiment results, GPT-4 and RF + GPT-4 show strong recall performance. Traditional machine learning models underperform due to problems with sparse sampling and data imbalance. GPT-4o and LGBM achieved high accuracy, likely due to predicting too many of the most common categories, but their effectiveness in identifying true positive cases was limited. DelayPTC-LLM demonstrated the most balanced results across all evaluation metrics, making it potentially the best choice, especially in applications that need to balance accuracy and recall.

Overall, the DelayPTC-LLM model demonstrated excellent performance on various performance metrics, notably maintaining high accuracy (0.83) while achieving satisfactory recall (0.50) and precision (0.66). This balanced performance indicates that DelayPTC-LLM is capable of effectively identifying true positive cases while reducing false positives, ensuring both the reliability and practicality of the predictions. Its relatively high F1-score further highlights this, showcasing the model's strong capability to balance recall and precision effectively. Such efficient and precise predictive ability positions DelayPTC-LLM as an ideal choice for complex data analysis tasks, particularly in applications requiring precise decision support.

**TABLE3 Performance Comparison of DelayPTC-LLM with Baseline Models**

| Model | Accuracy | Recall | Precision | F1-score |
|---|---|---|---|---|
| GPT-4 | 0.53 | 0.86 | 0.52 | 0.65 |
| GPT-4o | 0.80 | 0.10 | 0.23 | 0.12 |
| RF | 0.79 | 0.11 | 0.25 | 0.14 |
| LGBM | 0.82 | 0.02 | 0.35 | 0.04 |
| RF +GPT-4 | 0.55 | 0.76 | 0.24 | 0.37 |
| LGBM +GPT-4 | 0.68 | 0.52 | 0.27 | 0.36 |
| DelayPTC-LLM | 0.83 | 0.50 | 0.66 | 0.46 |

The results demonstrate that DelayPTC-LLM outperforms the other models in terms of predictive accuracy. The implementation of prompt engineering further enhances the performance of the DelayPTC-LLM, offering higher accuracy and better interpretability compared to traditional methods. This innovative approach allows for a more nuanced analysis of passenger decisions, reflecting a significant step forward in the application of LLMs to real-world transportation challenges. Moreover, the comparative analysis with traditional machine learning models such as RF and LGBM underscores the superior capability of DelayPTC-LLM in handling sparse and complex datasets typical of urban metro systems. The results not only affirm the superiority of DelayPTC-LLM over conventional models but also pave the way for future advancements in the field of intelligent transportation systems.

**Discussion**

The possibility of leveraging LLMs for passenger travel choice prediction presents a promising avenue. As shown in DelayPTC-LLM, employing prompt engineering and the CoT strategy significantly improves the predictive performance of LLMs, effectively addressing their limitations in prediction precision. By training LLMs on the travel choices dataset, they can uncover underlying preferences and trends that influence passenger decisions, such as choices between different modes of transport, route preferences, or the selection of travel periods.

Besides being trained on refined datasets of passenger travel choices, LLMs are also capable of handling multimodal data. With their sophisticated abilities to comprehend and produce text that resembles human communication, LLMs can effectively interpret and analyze a wide range of unstructured data, including customer feedback, travel patterns, and discussions on social media. Furthermore, LLMs can integrate with real-time data streams to provide dynamic predictions that reflect current travel conditions, thereby enabling more responsive and adaptive transportation services. For instance, during peak times, LLMs can harness real-time information from social media to predict shifts in passenger behavior. This capability enables the metro operation management to manage resources more effectively and provide personalized travel recommendations to passengers.

LLMs have shown significant potential in data mining and pattern recognition for predicting passenger travel choices. However, to improve predictive accuracy, it is essential to support the LLM with instructions, including prompt adjustments and other strategies. Additionally, the opaque nature of LLMs often makes their decision-making process challenging to understand, potentially leading to inexplicable predictions. Consequently, in practical applications, guiding the analysis and modeling processes with CoT is crucial to ensure that LLMs predict passenger travel behavior more accurately and

*Chen Chen, Yuxin He\*, Hao Wang, Jingjing Chen and Qin Luo*

effectively. Besides, the application of LLMs in this context also requires careful consideration of data privacy and the ethical implications of predictive analytics. Ensuring the security and anonymity of passenger data is paramount to maintaining trust and complying with regulations.

Overall, the use of LLMs for passenger choice prediction holds significant potential to transform public transport systems into more user-centric and efficient networks. With ongoing advancements in AI, coupled with robust data governance, LLMs could soon become an integral part of strategic planning and operational management in transportation.

## CONCLUSIONS

In conclusion, this research developed a data-driven framework for modeling passenger travel choices during metro delays, with an innovative application of DelayPTC-LLM in the predictive modeling component. This research effectively demonstrates the capabilities and advantages of using DelayPTC-LLM to predict passenger travel choices during metro delays. DelayPTC-LLM integrates delay logs and passenger travel choice datasets, and it has implemented CoT prompt engineering specifically for transportation big data. Furthermore, this research established baseline experiments using pure LLMs and conventional machine learning models for comparison. Benefiting from the LLM's superior spare data processing capability and the specifically designed CoT prompt engineering, this model's predictive performance significantly surpasses that of traditional machine learning models and standard LLMs. This research not only confirms the effectiveness of LLMs in a novel application domain but also paves the way for researches aimed at optimizing public transport system and improving URT service quality.

Overall, the research offers robust theoretical support for urban rail transportation and delay coping plans, leading to improved metro management and an enhanced passenger travel experience. The DelayPTC-LLM holds significant importance in mitigating the adverse effects of unexpected service disruptions, easing passenger flow congestion, and enhancing the quality of rail transit operation services.

## ACKNOWLEDGMENTS

This study was supported by the National Natural Science Foundation of China (72301180); Guangdong Basic and Applied Basic Research Foundation (2021A1515110731); Shenzhen Science and Technology Program (RCBS20231211090512002); Shanghai Key Laboratory of Rail Infrastructure Durability and System Safety (R202203); the Natural Science Foundation of Top Talent of SZTU (GDRC202126)

## AUTHOR CONTRIBUTIONS

The authors confirm their contribution to the paper as follows: study conception and design: C. Chen, Y. He; data collection: Y. He, Q. Luo; analysis and interpretation of results: C. Chen, Y. He., H. Wang; draft manuscript preparation: C. Chen, Y. He., J. Chen. All authors reviewed the results and approved the final version of the manuscript.